\documentclass[conference]{IEEEtran}
\IEEEoverridecommandlockouts
\usepackage{cite}
\usepackage{amsmath,amssymb,amsfonts}
\usepackage{algorithmic}
\usepackage{graphicx}
\usepackage{textcomp}
\usepackage{xcolor}
\def\BibTeX{{\rm B\kern-.05em{\sc i\kern-.025em b}\kern-.08em
    T\kern-.1667em\lower.7ex\hbox{E}\kern-.125emX}}

\usepackage{multirow}
\usepackage{threeparttable}
\usepackage{cite}
\usepackage[caption=false]{subfig}
\newcommand{\ineq}[1]{\footnotesize$#1$\normalsize}{}
\renewcommand{\baselinestretch}{0.87}

\begin{document}
\bstctlcite{IEEEexample:BSTcontrol}

\title{
A Case for Lifetime Reliability-Aware Neuromorphic Computing
}

\author{\IEEEauthorblockN{Shihao Song and Anup Das}
\IEEEauthorblockA{{Electrical and Computer Engineering}
{Drexel University}, Philadelphia, PA, USA\\
Email: \{shihao.song,anup.das\}@drexel.edu}
}

\maketitle

\begin{abstract}
Neuromorphic computing with non-volatile memory (NVM) can significantly improve performance and lower energy consumption of machine learning tasks implemented using spike-based computations and bio-inspired learning algorithms. High voltages required to operate certain NVMs such as phase-change memory (PCM) can accelerate aging in a neuron's CMOS circuit, thereby reducing the lifetime of neuromorphic hardware. In this work, we evaluate the long-term, i.e., lifetime reliability impact of executing state-of-the-art machine learning tasks on a neuromorphic hardware, considering failure models such as negative bias temperature instability (NBTI) and time-dependent dielectric breakdown (TDDB). Based on such formulation, we show the reliability-performance trade-off obtained due to periodic relaxation of neuromorphic circuits, i.e., a stop-and-go style of neuromorphic computing.
\end{abstract}


\section{Introduction}
Spiking neural network (SNN)~\cite{maass1997networks} is a machine learning technique designed using spike-based computation and bio-inspired learning algorithms~\cite{dan2004spike}. Neuromorphic hardware such as DYNAP-SE~\cite{moradi2017scalable}, TrueNorth~\cite{debole2019truenorth}, and Loihi~\cite{davies2018loihi} can execute SNN-based machine learning tasks in an energy-efficient manner, 
thanks to low-power neuron circuits~\cite{indiveri2003low}, distributed implementation of computing and storage as crossbars~\cite{merolla2011digital},
and the integration of non-volatile memory (NVM) for synaptic storage~\cite{burr2017neuromorphic,Mallik2017}. 
Several techniques are recently proposed to map and execute SNNs on to neuromorphic hardware \cite{song2020compiling,balaji20pycarl,balaji2019mapping,das2018mapping,das2018dataflow,balaji2020run}. These techniques mostly target performance (e.g., accuracy) and energy of neuromorphic computing. Unfortunately, neuromorphic hardware are prone to reliability issues such as limited programming endurance, read disturbance of NVM cells, and aging of CMOS-based neuron circuits~\cite{chen2018reliability,gleixner2009reliability,pirovano2004reliability}. In this work, we focus on the circuit aging due to negative bias temperature instability (NBTI) and time-dependent dielectric breakdown (TDDB) failure mechanisms~\cite{hu1993future,DasCASES2013,mneme}.


Due to the high voltage operating requirement of NVM, CMOS devices in a neuron circuit are exposed to high-voltage induced stress when propagating excitation (i.e., current) through an NVM synapse. This impacts the long-term, i.e., lifetime reliability of neuromorphic hardware.
As memory process technology scales down to smaller dimensions, reliability issues are expected to exacerbate due to the following three reasons. First, 
the electric field and power density increase in scaled nodes, exceeding their corresponding maximum value for reliable operation. Second, increasing power density also leads to higher chip temperatures and consequently, an even faster acceleration of the degradation mechanisms.
Third, new materials like high-\ineq{k} dielectrics and novel devices such as multi-gate field-effect transistor (FET) that are commonly used for the neuron circuit in neuromorphic hardware have unknown reliability behavior and they introduce new failure mechanisms at scaled nodes.
In our recent work \cite{balaji2019framework}, we have analyzed NBTI failure in neuromorphic computing. This work extends our earlier work in the following three directions. First, we consider other failure mechanisms such as TDDB and show the impact of system-level design decisions on the circuit aging in neuromorphic hardware. Second, we consider aging in a neuron circuit, which drives current into a crossbar to read synaptic weights stored in its NVM cells. Third, we show the performance-reliability trade-off in periodic relaxation of neuron excitations in neuromorphic hardware using state-of-the-art machine learning applications.



\section{Modeling Reliability of Crossbars}\label{sec:reliability}

\subsection{NBTI Issues in Neuromorphic Computing}
This is a failure mechanism of CMOS devices inside a neuron, when positive charges are trapped at the oxide-semiconductor boundary underneath the gate of a CMOS~\cite{gao2017nbti}.
NBTI manifests as 1) decrease in drain current and transconductance, and 2) increase in off current and threshold voltage. 
The lifetime of a CMOS device is measured in terms of its \textit{mean time to failure} (\textbf{MTTF}) as \ineq{\text{MTTF}_\text{NBTI} = \frac{A}{V^\gamma}e^{\frac{E_a}{KT}}},
where \ineq{A} and \ineq{\gamma} are material-related constants, \ineq{E_a} is the activation energy, \ineq{K} is the Boltzmann constant, \ineq{T} is the temperature, and \ineq{V} is the overdrive gate voltage of the CMOS device.

\begin{figure}[h!]
 	\centering
    \vspace{-6pt}
 	\centerline{\includegraphics[width=0.5\columnwidth]{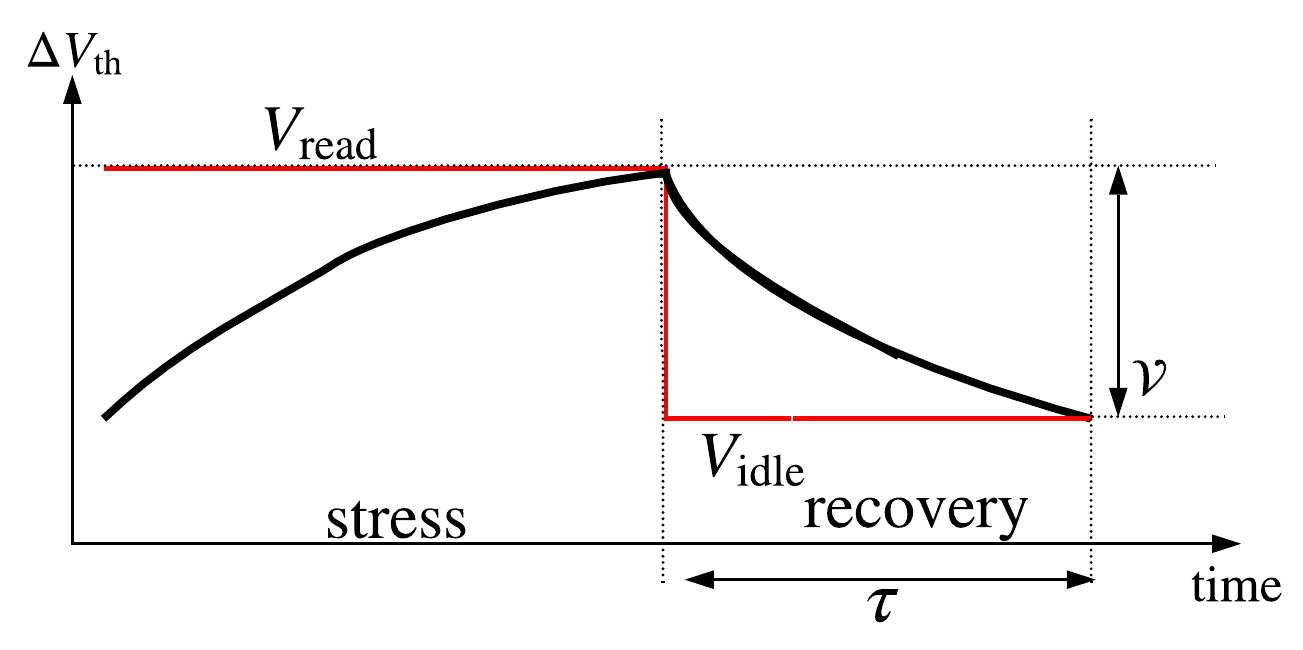}}
 	\caption{Demonstration of degradation due to NBTI.}
    \vspace{-10pt}
 	\label{fig:nbti_demo}
\end{figure}

Recent studies suggest that a portion of the threshold voltage can be recovered by annealing at high temperatures if the NBTI stress voltage is removed. 
Figure~\ref{fig:nbti_demo} illustrates the stress and recovery of threshold voltage of a CMOS device due to NBTI failure mechanism on application of a high (\ineq{V_\text{read}} = 1.8V) and a low voltage (\ineq{V_\text{idle}} = 1.2V) to a CMOS device in a neuron circuit. We observe that both stress and recovery depends on the time of exposure to the corresponding voltage~\cite{song2020improving}.

\vspace{-10pt}

\subsection{TDDB Issues in Neuromorphic Computing}
This is a failure mechanism in a CMOS device, when the gate oxide breaks down as a result of long-time application of relatively low electric field (as opposed to immediate breakdown, which is caused by strong electric field)~\cite{roussel2018new}. 
The TDDB lifetime of a CMOS device is \ineq{\text{MTTF}_\text{TDDB} = A.e^{-\gamma\sqrt{V}}},
where \ineq{A} and \ineq{\gamma} are material-related constants, and \ineq{V} is the overdrive gate voltage of the CMOS device~\cite{Das2014}. 



\subsection{Circuit Aging in Neuromorphic Computing}
\begin{figure}[htbp]%
	\centering
	\subfloat[][Spike times of a neuron in LeNet CNN. \label{fig:spike}]{
		\includegraphics[width=0.99\columnwidth]{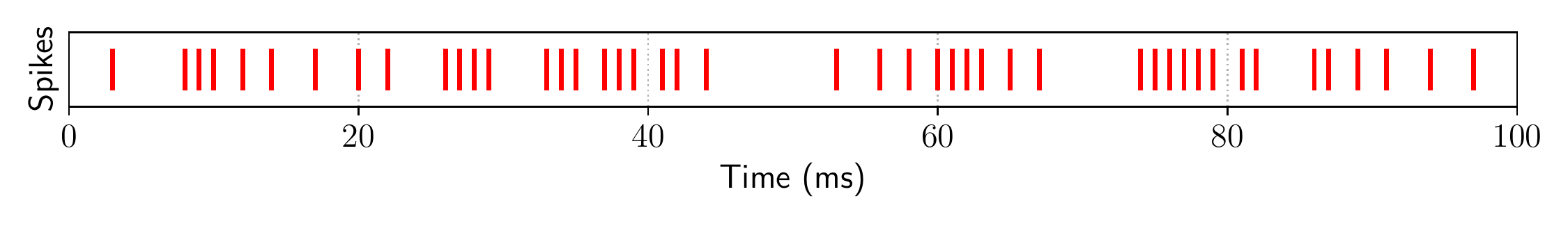}
	}
	\quad
	\subfloat[][Voltage of the neuron to process the spike train. \label{fig:voltage}]{
		\includegraphics[width=0.99\columnwidth]{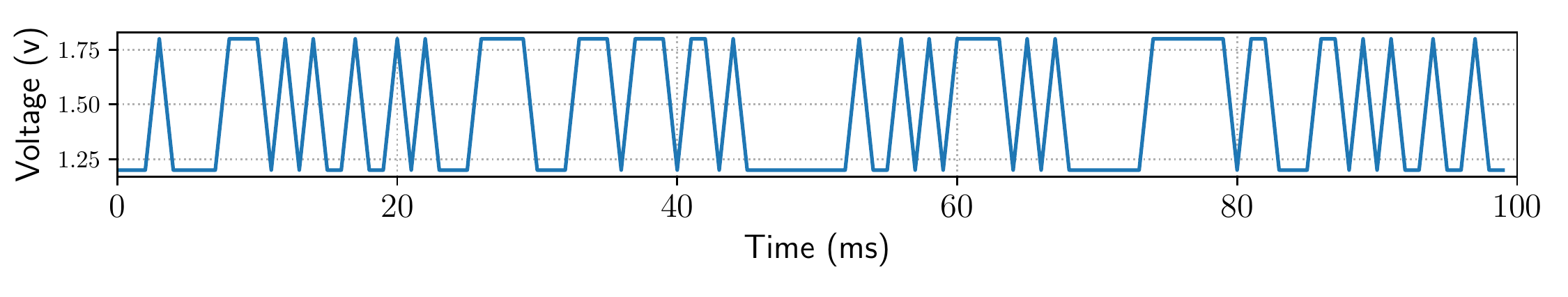}
	}
	\quad
	\subfloat[][NBTI aging (in log scale) of the neuron. \label{fig:nbti}]{
		\includegraphics[width=0.99\columnwidth]{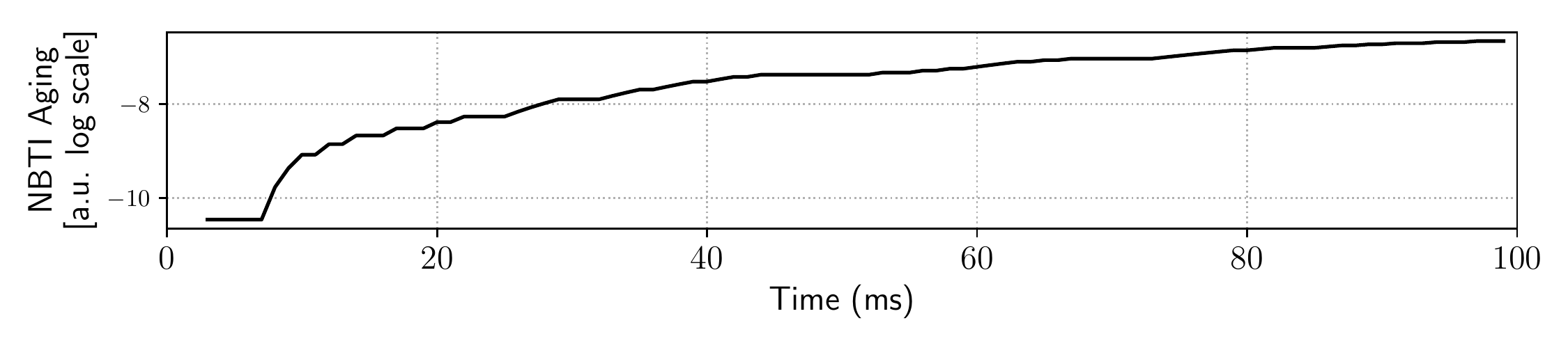}
	}
	\quad
	\subfloat[][TDDB aging (in log scale) of the neuron. \label{fig:tddb}]{
		\includegraphics[width=0.99\columnwidth]{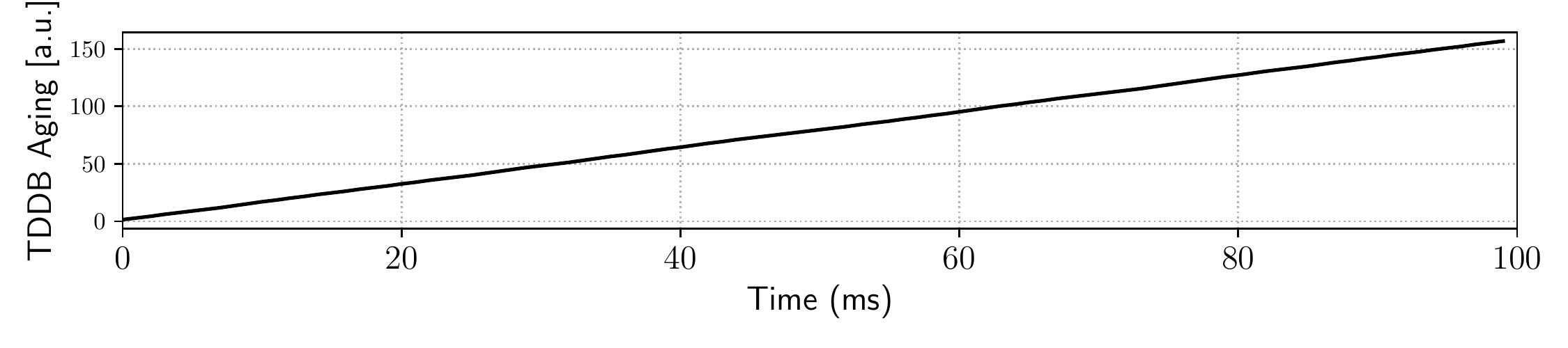}
	}
	\caption{(a) Spike times of a neuron in LeNet, (b) voltages needed to propagate these spikes through its fanout synapses, (c) NBTI degradation (in arbitrary units), and (d) TDDB degradation (in arbitrary units) of CMOS devices.}
	\label{fig:nbti_tddb_demo}
\end{figure}

To illustrate the degradation caused by these failure mechanisms, we take the example of a single neuron of the LeNet convolutional neural network (CNN) \cite{lecun2015lenet} used for handwritten digit recognition and illustrate its spike times within the first 100ms in Figure~\ref{fig:spike}. The voltage required to propagate these spikes through the neuron's fanout synapses are shown in Figure~\ref{fig:voltage}. Figures \ref{fig:nbti} and \ref{fig:tddb} show the NBTI and TDDB aging of a CMOS device inside the neuron's circuit, respectively. As can be clearly seen, both aging increases with time as more spikes are generated by the neuron. If CMOS devices in the neuron circuit are not de-stressed regularly, the aging (both NBTI and TDDB) in a neuron continues to increase, eventually leading to transient, intermittent, or permanent faults in the neuromorphic hardware.

\vspace{-10pt}
\vspace{-10pt}

To de-stress a neuron, all CMOS devices in the neuron must be programmed with a voltage lower than the threshold voltage \ineq{V_\text{th}}, which forces them to operate in the sub-threshold region, relieving their stress. Once discharged, a
neuron requires several clock cycles to boost its voltage
back to the required voltage level, before it can safely be used to generate spikes again. This introduces performance overhead.

\section{Periodic Relaxation of Neuromorphic Circuits}\label{sec:relaxation}
To improve the long-term, i.e., the lifetime reliability of neuromorphic computing, we propose periodic relaxation of a neuromorphic architecture, where we de-stress all neurons in the hardware at fixed intervals.
To compute the overhead due to such de-stress operations, we assume that the controller
issues a de-stress command to a crossbar once every \ineq{tDSI}, which is known as the \textit{de-stress interval}. Each de-stress operation completes within a
time interval \ineq{tDSC}, known as the \textit{de-stress cycle time}. 
Hence, the performance overhead (i.e., spike throughput loss) due to periodic de-stress is 
\begin{equation}
    \label{eq:destress_overhead}
    \footnotesize \text{de-stress overhead} = tDSC/tDSI.
    \vspace{-5pt}
\end{equation}

\begin{figure}[h!]
	\centering
	\vspace{-6pt}
	\centerline{\includegraphics[width=0.99\columnwidth]{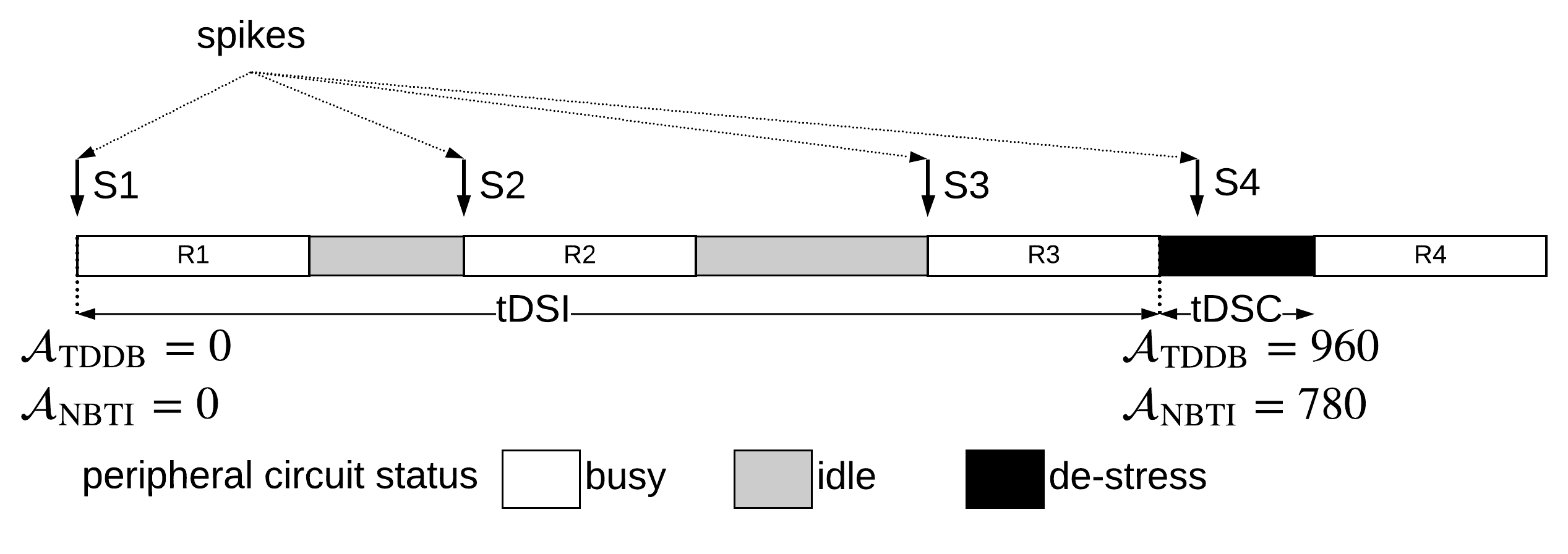}}
	\vspace{-10pt}
	\caption{Performance impact due to periodic relaxation.}
	\vspace{-10pt}
	\label{fig:motivation}
\end{figure}

Figure~\ref{fig:motivation} shows an example where four spikes (S1, S2, S3, \& S4) generated by a neuron. These spikes have some idle time between them. The neuron circuits are de-stressed after every \ineq{tDSI}, such that the aging due to NBTI and TDDB (indicated by \ineq{\mathcal{A}_\text{TDDB}} and \ineq{\mathcal{A}_\text{NBTI}}, respectively) are lower than 1000 units. Using this approach, the de-stress operation is initiated upon generating S3, which increases the latency of S4 due to the non-zero latency of the de-stress operation (indicated by \ineq{tDSC}). Increase in spike latency can lead to information loss in SNNs and degrade the quality of response.

\vspace{-10pt}

We introduce two key performance metrics in SNNs that are affected due to periodic de-stressing of neuromorphic architectures -- inter-spike interval (ISI) and disorder spike count. These are defined as follows.
\begin{itemize}
    \item \textbf{Inter-spike interval distortion:} Performance of supervised machine learning is measured in terms of \textit{accuracy}, which can be assessed from inter-spike intervals (ISIs)~\cite{grun2010analysis}. To define ISI, we let \ineq{\{t_1,t_2,\cdots,t_{K}\}} be a neuron's firing times in the time interval \ineq{[0,T]}. 
    The average ISI of this spike train is given by \cite{grun2010analysis}:
    \begin{equation}
        \vspace{-5pt}
        \label{eq:isi}
        \footnotesize \mathcal{I} = \sum_{i=2}^K (t_i - t_{i-1})/(K-1).
    \end{equation} 
    \item \textbf{Disorder spike count:} This is defined for SNNs where information is encoded in terms of spike rate.
    We formulate spike disorder as follows. Let \ineq{F^i = \{F_1^i,\cdots,F_{n_i}^i\}} be the expected spike arrival rate at neuron \ineq{i}  and \ineq{\hat{F}^i = \{\hat{F}_1^i,\cdots,\hat{F}_{n_i}^i\}} be the actual spike rate considering de-stress latencies. The spike disorder is computed as 
    \begin{equation}
    \label{eq:spike_disorder}
    \footnotesize \text{spike disorder} = \sum_{j=1}^{n_i} [(F_j^i - \hat{F}_j^i)^2] / n_i
    \end{equation}
\end{itemize}

\section{Evaluation}\label{sec:evaluation}
We evaluate 10 standard machine learning applications, which are listed in~Table \ref{tab:apps}.

\vspace{-15pt}
\begin{table}[h!]
    \caption{Applications used to evaluate our approach \cite{song2020compiling}.}
	\label{tab:apps}
	\renewcommand{\arraystretch}{0.8}
	\setlength{\tabcolsep}{2pt}
	\centering
	\begin{threeparttable}
	{\fontsize{6}{10}\selectfont
		\begin{tabular}{cc|ccl|c}
			\hline
			\textbf{Class} & \textbf{Applications} & \textbf{Synapses} & \textbf{Neurons} & \textbf{Topology} & \textbf{Accuracy}\\
			\hline
			\multirow{3}{*}{MLP} & EdgeDet & 272,628 &  1,372 & FeedForward (4096, 1024, 1024, 1024) & 100\%\\
			& ImgSmooth & 136,314 & 980 & FeedForward (4096, 1024) & 100\%\\
			& MLP-MNIST  & 79,400 & 984 & FeedForward (784, 100, 10) & 95.5\%\\
			\hline
			\multirow{4}{*}{CNN} & CNN-MNIST  & 159,553 & 5,576 & CNN & 96.7\%\\
			& LeNet-MNIST & 1,029,286 & 4,634 & CNN & 99.1\%\\
			& LeNet-CIFAR  & 2,136,560 & 18,472 & CNN & 84.0\%\\
			& HeartClass~\cite{HeartClassJolpe,Das2019} & 2,396,521 & 24,732 & CNN & 85.12\%\\
			\hline
 			\multirow{3}{*}{RNN} & HeartEstm~\cite{HeartEstmNN}  & 636,578 & 6,952 & Recurrent Reservoir & 99.2\%\\
 			& SpeechRecog  & 636,578 & 6,952 & Recurrent Reservoir & 96.8\%\\
 			& VisualPursuit  & 636,578 & 6,952 & Recurrent Reservoir & 89.0\%\\
			\hline
	\end{tabular}}
	\end{threeparttable}
\end{table}
\vspace{-15pt}

\subsection{Reliability}\label{sec:reliability_results}
\begin{figure}[h]%
	\centering
	\subfloat[][NBTI aging for 10 machine learning applications. \label{fig:nbti_summary}]{
		\includegraphics[width=0.99\columnwidth]{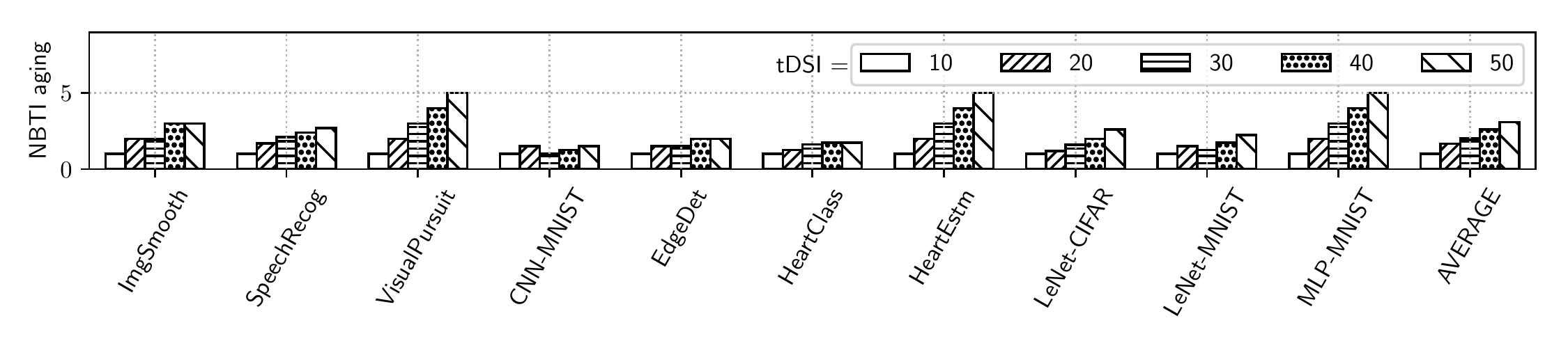}
	}
	\quad
	\subfloat[][TDDB aging for 10 machine learning applications. \label{fig:tddb_summary}]{
		\includegraphics[width=0.99\columnwidth]{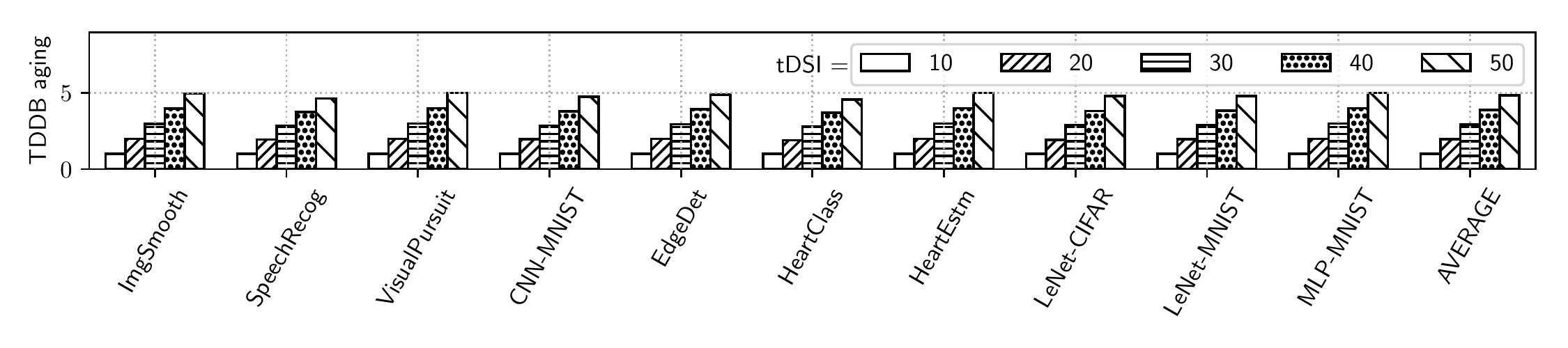}
	}
	\caption{(a) Normalized NBTI aging, and (b) Normalized TDDB aging for tDSI of 10ms, 20ms, 30ms, 40ms, and 50ms.}
	\label{fig:nbti_tddb_results}
\end{figure}

Figures \ref{fig:nbti_summary} and \ref{fig:tddb_summary} plot respectively, the NBTI and TDDB aging of the 10 machine learning applications when increasing the tDSI from 10ms to 50ms. We make the following three key observations. First, both NBTI and TDDB aging increases with increase in tDSI. This is because, a neuron accrues higher aging when its CMOS devices are kept active for longer duration (i.e., for higher tDSI). Second, the increase in aging is application dependent. For CNN-MNIST, increasing tDSI from 10ms to 50ms leads to 50\% increase in NBTI aging, compared to VisualPursuit, where the NBTI aging increase by 5x. This is because, the number of spikes generated in CNN-MNIST is far fewer than in VisualPursuit, which leads to lower aging in neuron circuits. Therefore, the impact of increasing tDSI for CNN-MNIST is less significant compared to VisualPursuit. 
Third, compared to NBTI, the increase of TDDB agings are consistent across different applications for the same range of tDSI. This is due to the difference in the two mechanisms. NBTI-induced stress (e.g., \ineq{V_\text{th}} shift) recovers partially when the neuron is idle. On the other hand, a CMOS devices encounters low-voltage TDDB stress even when idle.

\vspace{-10pt}
\vspace{-10pt}

\subsection{Performance}
\label{sec:performance_results}
\begin{figure}[h]%
	\centering
	\subfloat[][ISI distortion for 10 machine learning applications. \label{fig:isi_summary}]{
		\includegraphics[width=0.99\columnwidth]{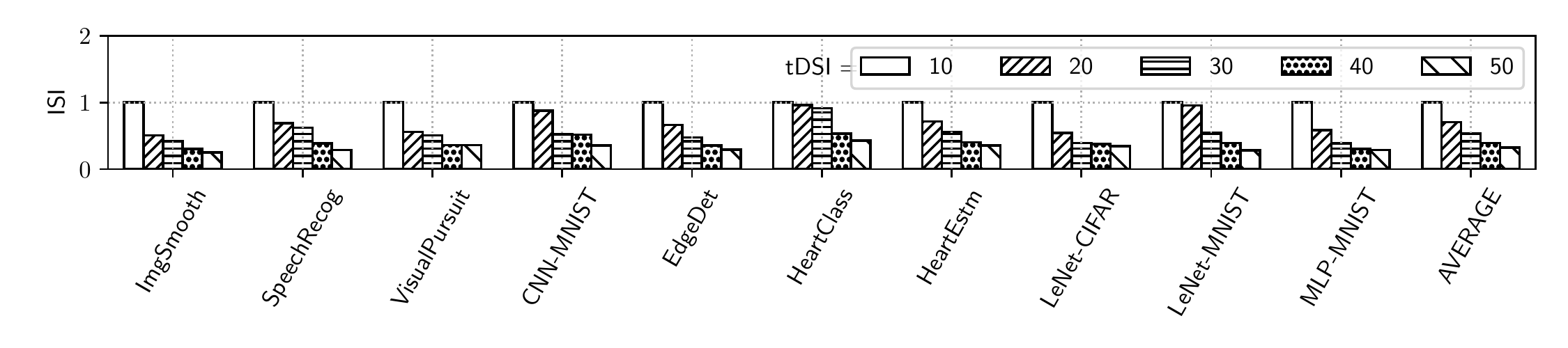}
	}
	\quad
	\subfloat[][Disorder spike count for 10 machine learning applications. \label{fig:dsc_summary}]{
		\includegraphics[width=0.99\columnwidth]{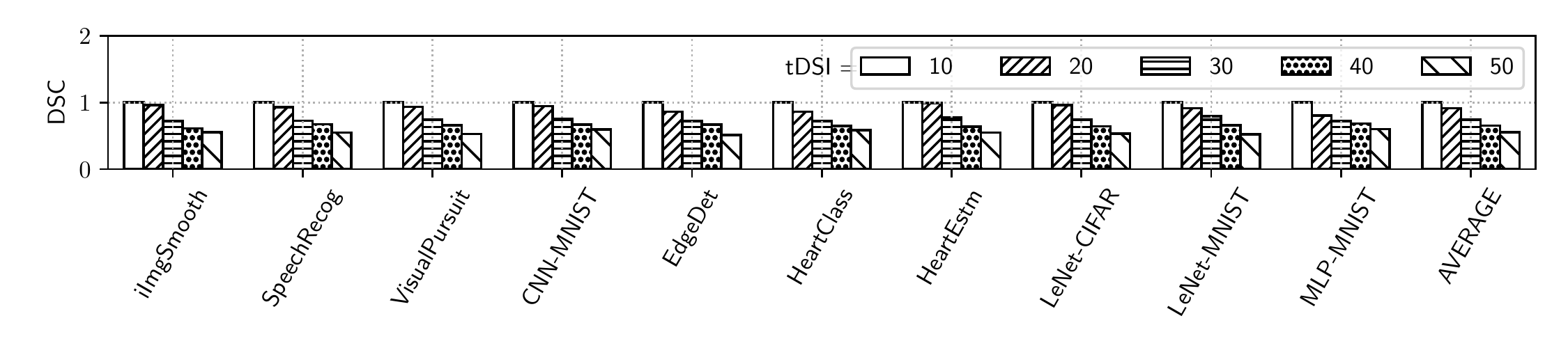}
	}
	\caption{ISI, disorder for tDSI of 10ms, 20ms, 30ms, 40ms, and 50ms.}
	\label{fig:isi_dsc_results}
\end{figure}

Figures \ref{fig:isi_summary} and \ref{fig:dsc_summary} plot respectively, the ISI distortion and disorder spike count (DSC) of the 10 machine learning applications when increasing the tDSI from 10ms to 50ms. We observe that both ISI and DSC reduces with increase in tDSI. This reduction is due to the reduction of the de-stress overhead (Equation~\ref{eq:destress_overhead}) with an increase in tDSI. 

\vspace{-10pt}

\subsection{Thermal Impact}
\begin{figure}[h!]
	\centering
	\vspace{-5pt}
	\centerline{\includegraphics[width=0.99\columnwidth]{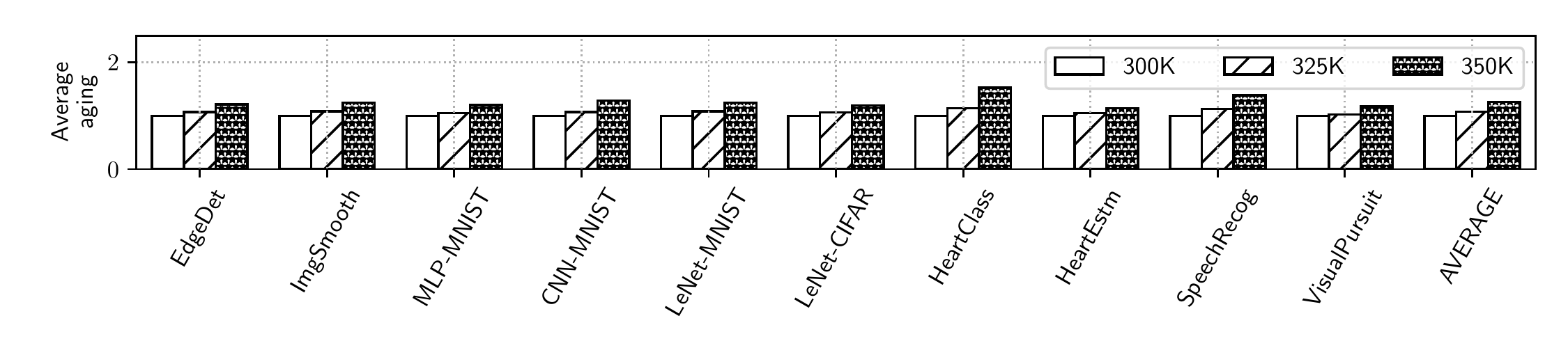}}
	\vspace{-10pt}
	\caption{Average circuit aging at 325K and 350K normalized to aging at 300K.}
	\vspace{-10pt}
	\label{fig:temperature}
\end{figure}

The results of Sections \ref{sec:reliability_results} are obtained at nominal temperature of 300K. Prior works such as \cite{DasTPDS16} show the impact of temperature on reliability of conventional multiprocessor system. Figure~\ref{fig:temperature} shows the increase of aging with temperature.
Average circuit aging at 325K and 350K is higher than that at 300K by an average of 7\% and 26\%, respectively.

\vspace{-10pt}

\section{Conclusion}\label{sec:conclusion}
We evaluate circuit aging in the neurons of neuromorphic architectures considering NBTI and TDDB failure mechanisms. We then propose a simple approach to improve reliability by periodically de-stressing its neurons. This introduces latency, which degrades key performance metrics such as inter-spike interval and disorder spike count, which correlates directly to the performance of machine learning models. We evaluate reliability-performance trade-offs for 10 state-of-the-art machine learning applications. We \textbf{conclude} that the proposed work will enable intelligent reliability optimization strategies in neuromorphic computing.

\section*{Acknowledgment}
This work is supported by the National Science Foundation Faculty Early Career Development Award CCF-1942697 (CAREER: Facilitating Dependable Neuromorphic Computing: Vision, Architecture, and Impact on Programmability).

\bibliographystyle{IEEEtran}
\bibliography{commands,neuromorphic}

\end{document}